\documentclass[letterpaper]{article} 
\usepackage{aaai2026}  
\usepackage{amssymb}
\usepackage{times}  
\usepackage{helvet}  
\usepackage{courier}  
\usepackage[hyphens]{url}  
\usepackage{graphicx} 
\usepackage{amsmath}
\usepackage{natbib}  
\usepackage{caption} 

\usepackage{algorithm}
\usepackage{algorithmic}
\usepackage{tabularx} 
\usepackage{multirow}
\usepackage{booktabs}
\usepackage{adjustbox} 
\usepackage{newfloat}
\usepackage{listings}
\DeclareCaptionStyle{ruled}{labelfont=normalfont,labelsep=colon,strut=off} 
\floatstyle{ruled}
\newfloat{listing}{tb}{lst}{}
\floatname{listing}{Listing}
\nocopyright 

\title{SWIFT:~A  General Sensitive Weight Identification Framework for Fast Sensor-Transfer Pansharpening{}}
\author{
    Zeyu Xia\equalcontrib,
    Chenxi Sun\equalcontrib,
    Tianyu Xin\equalcontrib,
    Yubo Zeng,
    Haoyu Chen,
    Liang-Jian Deng\thanks{Corresponding author.} 
}
\affiliations{
    \textsuperscript{\rm 1}University of Electronic Science and Technology of China\\

     2024040902024@std.uestc.edu.cn, 2024090902029@std.uestc.edu.cn,tyxin@std.uestc.edu.cn, liangjian.deng@uestc.edu.cn
}

\usepackage{bibentry}

\begin{document}

\maketitle

\begin{abstract}
Pansharpening aims to fuse high-resolution panchromatic (PAN) images with low-resolution multispectral (LRMS) images to generate high-resolution multispectral (HRMS) images. Although deep learning-based methods have achieved promising performance, they generally suffer from severe performance degradation when applied to data from unseen sensors. Adapting these models through full-scale retraining or designing more complex architectures is often prohibitively expensive and impractical for real-world deployment. To address this critical challenge, we propose a fast and general-purpose framework for cross-sensor adaptation, SWIFT (Sensitive Weight Identification for Fast Transfer). Specifically, SWIFT employs an unsupervised sampling strategy based on data manifold structures to balance sample selection while mitigating the bias of traditional Farthest Point Sampling, efficiently selecting only 3\% of the most informative samples from the target domain. This subset is then used to probe a source-domain pre-trained model by analyzing the gradient behavior of its parameters, allowing for the quick identification and subsequent update of only the weight subset most sensitive to the domain shift. As a plug-and-play framework, SWIFT can be applied to various existing pansharpening models. Extensive experiments demonstrate that SWIFT reduces the adaptation time from hours to approximately one minute on a single NVIDIA RTX 4090 GPU. The adapted models not only substantially outperform direct-transfer baselines but also achieve performance competitive with, and in some cases superior to, full retraining, establishing a new state-of-the-art on cross-sensor pansharpening tasks for the WorldView-2 and QuickBird datasets. 
\end{abstract}

\begin{links}
    \link{Code}{https://github.com/Zeyuxia/SWIFT}
\end{links}
\begin{figure}[t]
    \centering
    \includegraphics[width=0.90\linewidth]{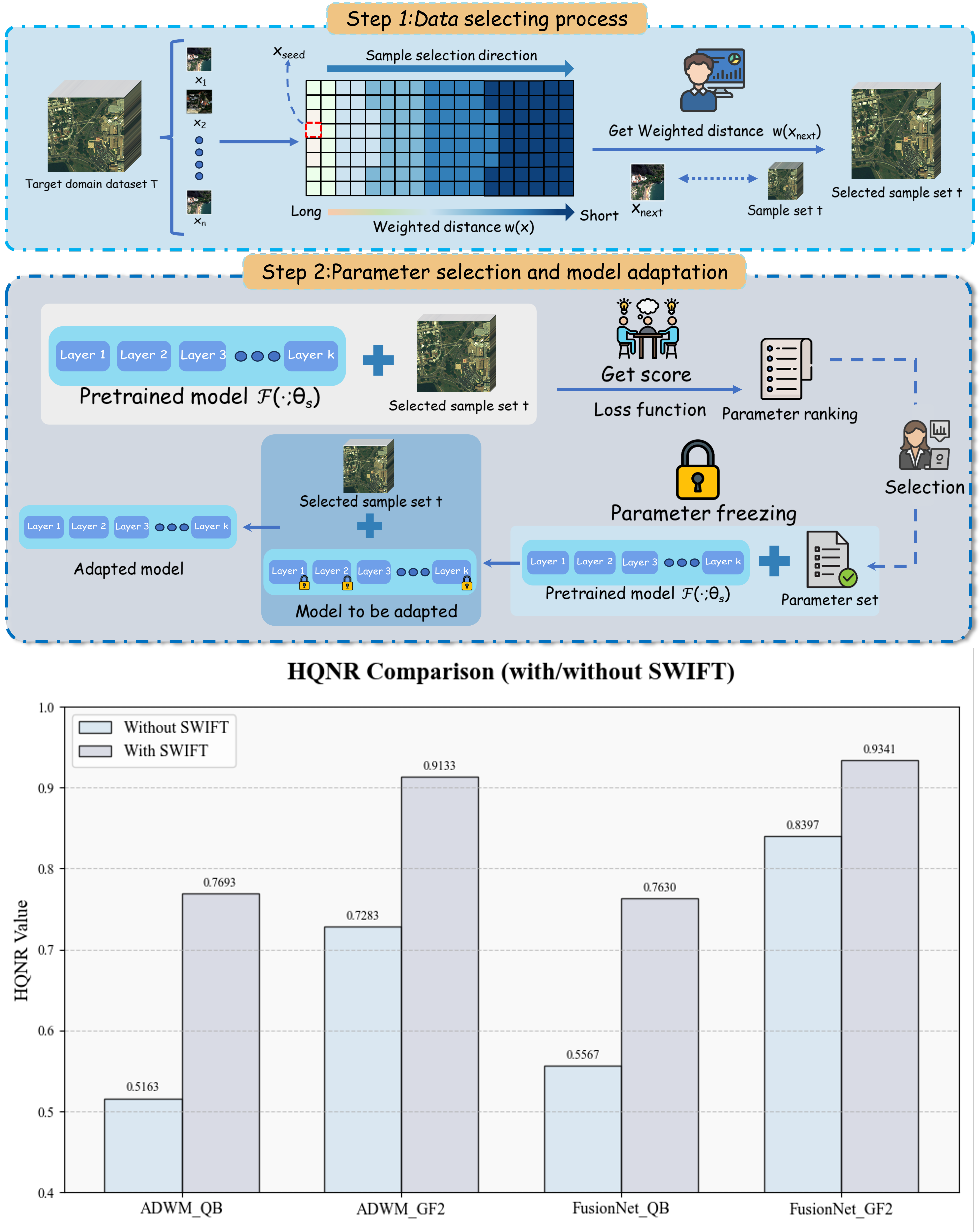}
    \caption{The SWIFT framework and its performance impact. Top: The two-step strategy of SWIFT. Bottom: HQNR comparison of representative models with and without SWIFT enhancement.}
    \label{fig:toutu}
\end{figure}

\section{Introduction} \label{sec:intro}

High-resolution multispectral (HRMS) images are essential for applications like urban planning and environmental monitoring \cite{Crampton2013,Fitzner2013}. However, due to  hardware limitations, satellites such as WorldView-2 (WV2) and Gaofen-2 (GF2) cannot capture images with both high spatial resolution and rich spectral information. To resolve this trade-off, these systems acquire two complementary data types: high-resolution panchromatic (PAN) and low-resolution multispectral (LRMS) images. The process of fusing these two kinds of images to generate high-resolution multispectral (HRMS) images is known as pansharpening.

Over the past few decades, pansharpening methods have evolved considerably, transitioning from traditional approaches to modern deep learning-based methods. Traditional approaches can be roughly categorized into three primary classes: component substitution (CS)\cite{choi2011,vivone2019}, multiresolution analysis (MRA)\cite{otazu2005,vivone2018}, and variational optimization (VO)\cite{wu2025,tian2021}. More recently, deep learning techniques have leveraged the powerful feature extraction capabilities of neural architectures. Models based on Convolutional Neural Networks (CNN) \cite{masi2016, he2019,Yang2023}, Transformer \cite{Li2024,wu2025,zhang2021,zhou2022} , and diffusion \cite{meng2023, zhong2024, rui2024} have already produced superior fusion results, significantly outperforming traditional methods.

Although deep learning methods excel at pansharpening, they generally suffer from severe \textit{performance degradation} when applied to target domain data that exhibit a distribution shift from the source domain. As satellite technology continues to diversify remote sensing data, this problem becomes increasingly critical. To address this \textit{cross-domain generalization} challenge, current strategies typically fall into a fundamental performance-cost dilemma. One line of work attempts to improve inherent robustness by designing more complex architectures built upon the original models; however, this strategy not only significantly increases model parameters and computational load, but also prevents simple, weight-only updates on deployed hardware, necessitating a full re-deployment cycle, and offers no guarantee of improved generalization, as demonstrated by the results of models like FusionNet\cite{quan2021fusionnet} and ADWM\cite{huang2025general} in our experiments. Another line of work relies on expensive retraining on the target domain or zero-shot methods. While effective, the substantial cost of time or computational resource is impractical in many application scenarios.

Based on the preceding analysis, two key challenges remain in enhancing generalization: (1) how to reduce the cost of computational resources and time, and (2) how to preserve the original model architecture. To our knowledge, no existing method can effectively address these two mutually constraining challenges simultaneously. 

Therefore, this paper introduces a novel task:how to improve a model's generalization capability at an extremely low computational and time cost, while preserving the original model architecture. To address this task, we propose a \textit{model-agnostic enhancement framework}—SWIFT. The core of this framework lies in a two-step ``targeted identification'' strategy as is shown in Figure \ref{fig:toutu}, encompassing data selection, followed by parameter selection and model adaptation.

Specifically, in the data selection step, we propose a Density-Aware Farthest Point Sampling (DA-FPS) strategy, which analyzes the data manifold to select a high-fidelity small subset of the target domain by balancing sample diversity from sparse regions with representativeness from dense regions. Subsequently, in the parameter selection step, we use this subset to identify the most sensitive parameters in the \textit{cross-domain task}. Inspired by network pruning \cite{LeCun1989}, our framework quantifies key gradient behaviors such as magnitude, consistency, and variance to calculate a composite sensitivity score for each parameter. We then select the most sensitive subset of parameters and update them, thereby achieving efficient and precise model adaptation.


In summary, the contributions of this study are as follows:

\begin{itemize}
    \item We are the first to define and address a novel task in the pansharpening field that enhances model generalization while jointly considering model architecture invariance and adaptation economy, offering a more practical research perspective for the community.
    
    \item We propose SWIFT, a plug-and-play and general-purpose framework. It innovatively integrates: (a) a data density-aware sample selection strategy, and (b) an efficient, gradient-based parameter sensitivity identification method, providing the first effective solution to the newly defined task.
    
    \item Extensive experiments demonstrate that SWIFT significantly improves cross-sensor performance of various mainstream models—achieving results comparable to or even surpassing full retraining on the target domain while requiring only minimal adaptation cost (approximately one minute, 3\% training data, and 30\% tunable parameters), and achieves state-of-the-art performance on multiple datasets, validating the effectiveness of our proposed framework.
\end{itemize}




\section{Related work}
\subsection{Deep Learning-based Pansharpening}
Deep learning has emerged as the dominant paradigm in pansharpening, based on network architecture, these methods can be broadly classified into three categories. Early CNN-based models, from the pioneering PNN \cite{masi2016} to advanced architectures like PanNet \cite{yang2017} and FusionNet \cite{deng2021}, excel at capturing local spatial-spectral patterns, but their limited receptive fields restrict the modeling of global dependencies, causing performance to degrade sharply under cross-sensor distribution shifts. To address this limitation, Transformer-based methods such as PanFormer \cite{zhou2022} incorporating self-attention to model long-range dependencies. Although this improves the accuracy of the fusion, the high computational cost limits their applicability in resource-constrained scenarios, and it is still necessary to deal with cross-domain generalization challenges. Finally, emerging hybrid methods, such as diffusion-based PanDiffusion \cite{meng2023}, aim to capture more complex data distributions.

However, these models are often tailored to specific sensors and require costly complete \textit{retraining} for cross-domain adaptation. Consequently, a core bottleneck for all deep learning models remains: a model trained in a source domain (e.g., QuickBird) often suffers from spectral distortion or spatial detail loss when applied to a target domain (e.g., GaoFen-2) due to distribution shifts caused by differing sensor characteristics.

\subsection{Existing Cross-Sensor Strategies of Pansharpening}

To address the adaptation challenges of deep learning models across different satellite sensors, existing research primarily follows three technical routes:

(1) \textbf{Model Retraining and Architectural Elaboration}: The most direct strategy to enhance performance in a target domain is to modify the model itself. This is mainly achieved in two ways. The first is model retraining, which involves conducting full end-to-end retraining of a source-domain pre-trained model on the target-domain dataset. This strategy not only requires a substantial amount of labeled target-domain data but also typically involves several hours of training time. The second is architectural elaboration, which aims to improve inherent robustness by designing more complex modules (e.g., additional attention layers or feature fusion units). Although this approach may effectively enhance generalization, it often significantly increases model parameters and computational load, and architectural changes prevent lightweight, weight-only updates on deployed hardware.

(2) \textbf{Parameter-Efficient Fine-Tuning}: Parameter-Efficient Fine-Tuning (PEFT) offers a paradigm for adapting large, general-purpose models to downstream tasks. In pansharpening, methods like PanAdapter \cite{wu2025} exemplify this by using new, trainable adapters to fine-tune pre-trained restoration models. In contrast, our framework is specifically designed for the cross-sensor adaptation of existing pansharpening models, rather than adapting a general restoration model to a new task. It achieves this by employing a data-driven analysis to identify and retune only the most critical original parameters.

(3) \textbf{Zero-Shot Methods}: These methods perform instance-specific optimization during inference using only a single input LRMS-PAN pair, thus requiring no additional training data. While they obviate the need for large-scale datasets, they typically require several minutes to process a single image and do not always achieve optimal fusion quality. This poses significant efficiency challenges for cross-sensor generalization and fails to meet the demands of rapid applications.

In summary, existing technical routes fail to strike an ideal balance among model architecture invariance, adaptation economy, and final performance. This underscores the necessity and urgency of the new task we propose— achieving generalization enhancement at an extremely low cost.
\begin{figure*}[ht]
  \centering
  \includegraphics[width=0.95\textwidth]{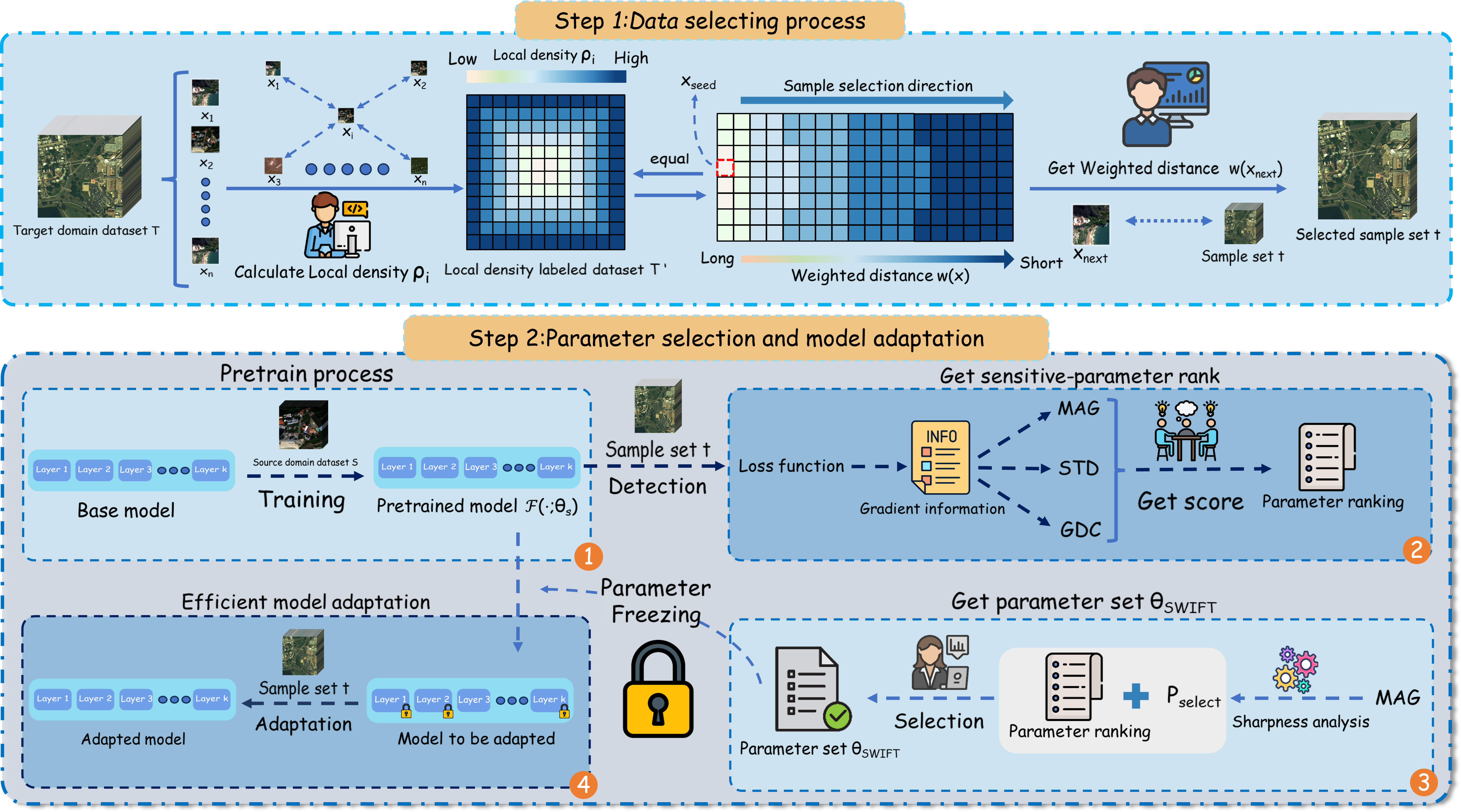}
  \caption{Overall Framework of the SWIFT framework }
  \label{overall_framework}
\end{figure*}

\section{Methodology}

\subsection{Overall framework} \label{sec:overall_framework}

The overall framework of our proposed method is illustrated in Figure\ref{overall_framework}. Let $\mathcal{F}(\cdot \, ; \theta_S)$ denote a given pansharpening model with parameters $ \theta_S$ 
  that has been pre-trained on a source domain. For the target domain, we are given a large set of unlabeled image pairs, denoted as $T = \{ Y_T, P_T \}$, where $Y_T$ and $P_T$ represent the low-resolution multispectral (LRMS) and high-resolution panchromatic (PAN) images, respectively. Our framework aims to address a novel task: to rapidly adapt the pre-trained model $\mathcal{F}(\cdot \, ; \theta_S)$ to the target domain at an extremely low computational cost without altering its architecture. To achieve this, the SWIFT framework performs adaptation through an efficient two-step ``targeted identification'' strategy.
  
First, in the data selection step, we employ a density-aware farthest point sampling strategy to efficiently select a small yet informative representative sample subset, $t$, from the large target-domain dataset T. This strategy is designed to preserve the core distributional features of the target domain using a minimal amount of data (e.g., 3\%). Subsequently, in the parameter selection and model adaptation step, we leverage the sample subset t to perform a parameter sensitivity analysis on the pre-trained model$\mathcal{F}(\cdot \, ; \theta_S)$
then we select and update only the subset of parameters $ \theta_{SWIFT} \subset\theta_S$, that is most sensitive to the target domain shift, thereby achieving rapid model adaptation. These two core stages will be detailed sequentially in subsequent sections.  
   
\subsection{Density-Aware Farthest Point Sampling Strategy} \label{sec:DA_FPS}

In model adaptation, the informational value of target-domain samples varies significantly: samples in dense regions are often redundant, while critical information in sparse regions can be easily overlooked. To address this problem, we introduce a \textit{ Density-Aware Farthest Point Sampling (DA-FPS)} strategy, which quantifies both density and coverage to select high-value samples, thereby constructing a high-fidelity microcosm of the target domain with a minimal subset. The process could be divided into two steps:

First, to quantify the ``uniqueness'' of each sample on the data manifold, we calculate the local density for each sample$x_i$ in the target-domain dataset $T = \{ x_1, x_2,\ldots,x_N \}$ using Kernel Density Estimation (KDE):
\begin{equation}
    \rho_i = \sum_{j \neq i} \exp \left( - \left( \frac{d(x_i, x_j)}{\sigma} \right)^2 \right)
\end{equation}
where $d(x_i,x_j)$ denotes the Euclidean distance between samples $x_i$ and $x_j$:
\begin{equation}
    d(x_i, x_j) = \sqrt{\sum_{k=1}^{D} (x_{i,k} - x_{j,k})^2}
\end{equation}
and $\sigma$ is a hyperparameter controlling the kernel bandwidth. A larger $\rho_i$ indicates that sample$x_i$ has high similarity to other samples and is located in a dense region of the data manifold; conversely, a smaller $\rho_i$ suggests that the sample is in a sparse region, containing more valuable edge or rare information, then we will get a local density labeled dataset $T' = \{ x_1, x_2,\ldots,x_N \}$.

After obtaining the density information for all samples, we iteratively construct the ``essence'' sample set $t$. The process is initialized by selecting the sample with the lowest global density that expressed as $arg\min_i\rho_i$,as the initial seed to ensure that the rarest information is not missed. In each subsequent iteration, for every unselected sample $\ x_i \notin t $, we compute a weighted distance $w(x_i)$:
\begin{equation}
    w(x_i) = d_{min}(x_i,t)\times\left(1 - \alpha \cdot \frac{\rho_i}{\max_j \rho_j} \right)
\end{equation}
where $d_{min}(x_i,t) = min_t\|x_i - s\|_2$ represents the shortest distance from sample $x_i$to the currently selected set $t$,ensuring spatial coverage (globality). The weighting term, using the normalized density, adjusts the selection preference, with $\alpha$ serving as a balancing factor. This mechanism gives higher priority to lower-density samples (smaller $\rho_i$),which receive a larger weighted distance $w(x_i)$. In each round, the sample with the maximum weighted distance, $x_{\text{next}} = \operatorname{arg\,max}_{x_i \in T \setminus t} w(x_i)$,is added to $t$  until the predefined sampling ratio r (i.e.,$|t| = \lfloor r \cdot N \rfloor$) is reached.

\subsection{Sensitive Parameter Selection and Efficient Model Adaptation}

To efficiently adapt the pre-trained model to the target domain, after obtaining the high-information sample subset, we proceed with parameter selection and model updating. We posit that parameters more critical to the model adaptation task should exhibit a combination of high impact, high definiteness, and high stability in their gradient signals when exposed to target-domain data.  To quantify these characteristics, we further divide the ``essence'' sample subset $t$ into $M$ microbatches. For any given trainable parameter $\theta_j$ in the model,we perform a full forward and backward pass for each microbatch $i$ to obtain a set of gradients $G_j=\{g_{j,1}.g_{j,2},\dots,g_{j,M}\}$. Based on this set, we compute the following three core metrics:

First, we calculate the \textit{Gradient Magnitude (MAG)}, which quantifies the average impact of the parameter on the model output. A larger magnitude indicates a higher impact on the loss reduction. It is formulated as follows:follows:
\begin{equation}
    \operatorname{MAG}(\theta_j) = \frac{1}{M} \sum_{i=1}^{M} \left| \mathbb{E}[g_{j,i}] \right|
\end{equation}
where $\mathbb{E}[g_{j,i}]$ is the expectation of all gradient values for parameter $\theta_j$ within the $i$-th microbatch.

Second, we compute the \textit{Gradient Direction Consistency (GDC)}, which is defined as:
\begin{equation}
    \operatorname{GDC}(\theta_j) = \frac{1}{M} \sum_{i=1}^{M} \frac{\max(N_{\text{pos}}, N_{\text{neg}})}{N_{\text{pos}} + N_{\text{neg}}}
\end{equation}
where $N_{pos}$ and $N_{neg}$ are the counts of positive and negative gradients for parameter $\theta_j$ in the $i$-th microbatch, respectively. This metric measures the consistency of the expected update direction across different samples. A GDC value closer to 1 indicates a more definite update direction.

Finally, we calculate the \textit{Gradient Standard Deviation (STD)}, formulated as:
\begin{equation}
    \operatorname{STD}(\theta_j) = \sqrt{\frac{1}{M} \sum_{i=1}^{M} \operatorname{Var}(g_{j,i})}
\end{equation}
where $Var(g_{j,i})$ is the variance of the gradients for parameter $\theta_j$ in the $i$-th microbatch.This metric reflects the stability of the update intention; a lower STD signifies a more stable update process.

To synthesize the importance of these three metrics, we normalize them and devise a weighted sum to compute a composite sensitivity score for each trainable parameter $\theta_j$:
\begin{equation}
    S(\theta_j) = \alpha \cdot \overline{\operatorname{MAG}(\theta_j)} + \beta \cdot (1 - \overline{\operatorname{STD}(\theta_j)}) + \gamma \cdot \overline{\operatorname{GDC}(\theta_j)}
\end{equation}
where $\overline{\operatorname{MAG}}$,$\overline{\operatorname{STD}}$,and $\overline{\operatorname{GDC}}$ are the normalized metrics, and $\alpha$,$\beta$,$\gamma$ are hyperparameters controlling their respective importance,satisfying $\alpha + \beta + \gamma = 1$ .The specific values of these weights and their impact on performance will be discussed in the ablation studies part. Notably, we assign a negative weight to the standard deviation term to reward parameters with more stable gradients.

After obtaining and ranking the sensitivity scores for all parameters in descending order, we employ a dynamic thresholding mechanism for parameter selection. First, we quantify the sharpness of the gradient distribution:
\begin{equation}
    \mathcal{H} = \operatorname{std}(\{\overline{\operatorname{MAG}_j}\}) + (\max(\{\overline{\operatorname{MAG}_j}\}) - \operatorname{median}(\{\overline{\operatorname{MAG}_j}\}))
\end{equation}
we define parameter sharpness as the sum of the standard deviation and the skewness of the distribution, where $\{\overline{\text{MAG}}_j\}$ is the set of normalized average gradient magnitudes across all parameters. A higher $\mathcal{H}$ value signifies a sharper distribution of influence, indicating a clear distinction between important and unimportant parameters. Conversely, a lower value suggests a flatter distribution. Subsequently, based on the calculated sharpness $\mathcal{H}$, we dynamically determine the optimal selection ratio $P_{select}$ for the current adaptation task to identify the final parameter subset $\theta_{SWIFT}$:
\begin{equation}
    \label{eq:p_select_simplified}
    \mathbf{P}_{\text{select}} = \eta_{\min} + (\eta_{\max} - \eta_{\min}) \cdot \operatorname{clip}(\mathcal{H}_{\text{norm}}, 0, 1)
\end{equation}
where $\mathcal{H}_{\text{norm}}$ is defined as:
\begin{equation}
    \label{eq:h_norm}
    \mathcal{H}_{\text{norm}} = \frac{\mathcal{H}-\mathcal{H}_{\min}}{\mathcal{H}_{\max}-\mathcal{H}_{\min}}
\end{equation}
Finally, we freeze all trainable parameters outside the subset $\theta_{SWIFT}$ and update this sensitive subset using the sample subset t to achieve efficient model adaptation.

\section{Experiment}

\subsection{Datasets and Metrics}
\textbf{Datasets:} We investigate the effectiveness of the proposed method on a wide range of datasets, including an 8 band dataset from WorldView-3 (WV3) and WorldView-2 (WV2) sensors, and 4-band datasets from QuickBird (QB) and GaoFen-2 sensors. Notably, we leverage Wald’s protocol to simulate the source  data due to the unavailability of ground truth (GT) images. Taking WV3 as an instance, we use 10000 PAN/LRM S/GT image pairs (64 $\times$ 64) for network training. For testing, we take 20 PAN/LRMS/GT image pairs (256 $\times$ 256) for reduced-resolution evaluation, and 20 PAN/LRMS image pairs (512 $\times$ 512) for the full-resolution assessment, which lacks GT images.

\textbf{Metrics:}
The quality evaluation is conducted both at reduced and  full resolutions. For reduced resolution tests, the widely used  SAM \cite{Yuhas1992}, ERGAS \cite{Wald2002}, SCC \cite{zhou1998}, and Q-index  for 4-band (Q4) and 8-band data (Q8) \cite{5159503} are adopted to assess the quality of the results. To evaluate the performance at full resolution, the HQNR, the $D_\lambda$, and the $D_s$ \cite{Vivone2015} indexes are considered.

\subsection{Generalization Ability, Training Details and Benchmark}

\begin{table}[htbp]
  \centering 
  \setlength{\tabcolsep}{3pt} 
  \renewcommand\arraystretch{1.0}  
     \begin{tabular}{
       p{1.7cm} 
       >{\centering\arraybackslash}p{1.85cm} 
       >{\centering\arraybackslash}p{1.85cm} 
       >{\centering\arraybackslash}p{2.05cm} 
     }
      \toprule
      \toprule
      \multirow{2}{*}{Method} & \multicolumn{3}{c}{GF2: Avg$\pm$std} \\ 
      \cmidrule(lr){2-4}  
       & $D_\lambda\downarrow$ & $D_s\downarrow$ & HQNR$\uparrow$ \\
      \midrule
          ZS-Pan & \underline{0.036$\pm$0.016} & 0.071$\pm$0.019 & \underline{0.896$\pm$0.027} \\
          PSDip & 0.045$\pm$0.021 & 0.075$\pm$0.021 & 0.883$\pm$0.022 \\
          PanNet & 0.220$\pm$0.040 & 0.198$\pm$0.026 & 0.625$\pm$0.029 \\
          FusionNet & 0.111$\pm$0.050 & \underline{0.055$\pm$0.014} & 0.840$\pm$0.051 \\
          U2Net & 0.137$\pm$0.047 & 0.134$\pm$0.038 & 0.747$\pm$0.052 \\
          SSDiff & 0.144$\pm$0.075 & 0.060$\pm$0.041 & 0.805$\pm$0.084 \\
          ADWM & 0.076$\pm$0.019 & \textbf{0.039$\pm$0.013} & 0.888$\pm$0.030 \\
          WFANet & \textbf{0.035$\pm$0.012} & 0.066$\pm$0.019 & \textbf{0.900$\pm$0.012} \\ 
      \bottomrule
      \bottomrule
    \end{tabular}
  \caption{Average quantitative metrics and standard deviations of some deep learning methods that trained and test on the GF2 datasets.}
  \label{tab:single_qb_gf2_full_results}
\end{table}

\textbf{Generalization Task:}
A core bottleneck for deep learning-based pansharpening is cross-domain generalization. Specifically, while models perform excellently on source-domain data, their performance degrades significantly when applied to a target domain with a distribution shift, as shown in table\ref{tab:single_qb_gf2_full_results} and \ref{tab:qb_gf2_full_results}. Pre-trained model shift to target domain always have a huge performence degration compared to model trained on target domain directly. Based on this, enhancing the performance of source domain pre-trained models has become a key goal for improving cross-domain generalization ability.

\textbf{Training Details:}
To evaluate the cross-domain generalization, we denote our experimental settings as ``SourceDomain-TargetDomain'', where the model is pre-trained on the SourceDomain and adapted and tested on TargetDomain. For instance, \textit{GF2-QB} in our tables indicates a model pre-trained on GaoFen-2 and tested on QuickBird dataset. All experiments are conducted on an NVIDIA RTX 4090 GPU with 24GB of video memory.

\begin{table*}[!t]
\begin{minipage}{\linewidth}
  \centering 
  \setlength{\tabcolsep}{3pt}
  \renewcommand\arraystretch{1.1}
    \begin{tabular}{
      >{\raggedright\arraybackslash}p{1.9cm}
      >{\centering\arraybackslash}p{1.94cm} 
      >{\centering\arraybackslash}p{1.94cm} 
      >{\centering\arraybackslash}p{1.94cm} 
      >{\centering\arraybackslash}p{1.08cm}
      >{\centering\arraybackslash}p{1.94cm}
      >{\centering\arraybackslash}p{1.94cm}
      >{\centering\arraybackslash}p{1.94cm}
      >{\centering\arraybackslash}p{1.08cm}
    }
      \toprule
      \toprule
      \multirow{2}{*}{Method} & \multicolumn{4}{c|}{QB-GF2: Avg$\pm$std} & \multicolumn{4}{c}{GF2-QB: Avg$\pm$std} \\
      \cmidrule{2-5}\cmidrule{6-9}
       & $D_\lambda\downarrow$ & $D_s\downarrow$ & HQNR$\uparrow$ & Time(s) & $D_\lambda\downarrow$ & $D_s\downarrow$ & HQNR$\uparrow$ & Time(s)\\
      \midrule
          PanNet & 0.220$\pm$0.040 & 0.198$\pm$0.026 & 0.625$\pm$0.023 & 12870.6 & 0.285$\pm$0.045 & 0.232$\pm$0.0227 & 0.549$\pm$0.026 & 15486.4 \\
          our.PanNet & 0.090$\pm$0.019 & 0.056$\pm$0.012 & 0.860$\pm$0.026 & 1399.6 & 0.193$\pm$0.059 & 0.172$\pm$0.025 & 0.668$\pm$0.023 & 1630.4 \\
          \midrule 
          FusionNet & \underline{0.111$\pm$0.050} & \textbf{0.055$\pm$0.014} & \underline{0.840$\pm$0.051} & \textbf{5874.3} & 0.311$\pm$0.045 & 0.191$\pm$0.039 & 0.557$\pm$0.033 & \textbf{6435.2} \\
          our.FusionNet & 0.041$\pm$0.037 & \textbf{0.027$\pm$0.006} & \textbf{0.934$\pm$0.035} & \textbf{61.4} & 0.167$\pm$0.050 & 0.048$\pm$0.016 & 0.794$\pm$0.056 & \textbf{69.3} \\
          \midrule 
          U2Net & 0.137$\pm$0.047 & 0.134$\pm$0.038 & 0.747$\pm$0.052 & \underline{11023.3} & \underline{0.105$\pm$0.057} & \textbf{0.075$\pm$0.029} & \underline{0.828$\pm$0.054} & \underline{12769.2} \\
          our.U2Net & 0.033$\pm$0.020 & 0.071$\pm$0.013 & 0.898$\pm$0.017 & \underline{460.2} & 0.132$\pm$0.040 & \underline{0.041$\pm$0.031} & 0.834$\pm$0.059 & \underline{572.4} \\
          \midrule 
          SSDiff & 0.144$\pm$0.075 & \underline{0.060$\pm$0.041} & 0.805$\pm$0.084 & 32336.5 & \textbf{0.082$\pm$0.038} & \underline{0.076$\pm$0.025} & \textbf{0.850$\pm$0.047} & 35972.1 \\
          our.SSDiff & \underline{0.033$\pm$0.029} & \underline{0.036$\pm$0.015} & \underline{0.932$\pm$0.021} & 1077.9 & \textbf{0.033$\pm$0.012} & \textbf{0.026$\pm$0.016} & \textbf{0.942$\pm$0.026} & 1199.1 \\
          \midrule 
          ADWM & 0.161$\pm$0.046 & 0.133$\pm$0.035 & 0.728$\pm$0.060 & 13645.9 & 0.795$\pm$0.196 & 0.127$\pm$0.182 & 0.516$\pm$0.177 & 26284.6 \\
          our.ADWM & 0.049$\pm$0.034 & 0.040$\pm$0.015 & 0.913$\pm$0.032 & 1091.3 & 0.107$\pm$0.035 & 0.140$\pm$0.030 & 0.769$\pm$0.055 & 971.7 \\
          \midrule 
          WFANet & \textbf{0.035$\pm$0.012} & 0.066$\pm$0.019 & \textbf{0.900 $\pm$0.012} & 24994.5 & 0.304$\pm$0.082  & 0.242$\pm$0.044  & 0.526$\pm$0.059  & 26787.0 \\
          our.WFANet & \textbf{0.020$\pm$0.013}  & 0.061$\pm$0.010 & 0.920$\pm$0.012  & 1612.4 & \underline{0.074$\pm$0.022}  & 0.085$\pm$0.026 & \underline{0.849$\pm$0.042} & 1856.7 \\
      \bottomrule
      \bottomrule
    \end{tabular}
  \caption{Mean values and standard deviations of all comparative methods on 20 full-resolution samples from the QB and GF2 datasets. Best:\textbf{bold} , and second best: :\underline{underline}.}
  \label{tab:qb_gf2_full_results}
  \small
  \centering 
  \setlength{\tabcolsep}{2pt}
  \renewcommand\arraystretch{1.1}
    \begin{tabular}{
       >{\raggedright\arraybackslash}p{1.72cm}
      >{\centering\arraybackslash}p{1.86cm} 
      >{\centering\arraybackslash}p{1.86cm} 
      >{\centering\arraybackslash}p{1.805cm} 
      >{\centering\arraybackslash}p{1.805cm}
      >{\centering\arraybackslash}p{1.86cm}
      >{\centering\arraybackslash}p{1.86cm}
      >{\centering\arraybackslash}p{1.805cm}
      >{\centering\arraybackslash}p{1.805cm}
    }
      \toprule
      \toprule
      \multirow{2}{*}{Method} & \multicolumn{4}{c|}{QB-GF2: Avg$\pm$std} & \multicolumn{4}{c}{GF2-QB: Avg$\pm$std} \\
      \cmidrule{2-5}\cmidrule{6-9}
       & SAM$\downarrow$ & ERGAS$\downarrow$ & SCC$\uparrow$ & Q2N$\uparrow$ & SAM$\downarrow$ & ERGAS$\downarrow$ & SCC$\uparrow$ & Q2N$\uparrow$\\
      \midrule
          PanNet       & 14.229$\pm$2.830 & 16.524$\pm$2.339 & 0.647$\pm$0.010 & 0.254$\pm$0.030 & 18.993$\pm$3.919 & 16.949$\pm$1.797 & \underline{0.757$\pm$0.018} & 0.402$\pm$0.075 \\
          our.PanNet   & 4.560$\pm$0.771  & 4.746$\pm$1.019  & 0.762$\pm$0.022  & 0.286$\pm$0.030  & 15.395$\pm$3.037 & 15.069$\pm$1.630 & 0.655$\pm$0.025 & 0.315$\pm$0.050 \\
          \midrule 
          FusionNet    & 3.208$\pm$0.476  & 3.873$\pm$0.627  & 0.787$\pm$0.019  & 0.600$\pm$0.071  & 13.924$\pm$2.319 & 13.881$\pm$1.155 & 0.724$\pm$0.027 & 0.455$\pm$0.096 \\
          our.FusionNet& 1.875$\pm$0.259  & 1.991$\pm$0.299  & 0.921$\pm$0.013  & 0.849$\pm$0.036  & 8.064$\pm$1.563  & 6.939$\pm$0.468  & 0.922$\pm$0.016 & 0.730$\pm$0.111 \\
          \midrule 
          U2Net        & 3.571$\pm$0.406  & \textbf{2.865$\pm$0.354}  & \underline{0.909$\pm$0.010}  & \textbf{0.804$\pm$0.054}  & \underline{13.279$\pm$1.577} & \underline{12.243$\pm$1.078} & 0.756$\pm$0.030 & \underline{0.593$\pm$0.116} \\
          our.U2Net    & \underline{1.222$\pm$0.202}  & \underline{1.186$\pm$0.193}  & \underline{0.969$\pm$0.005}  & \underline{0.946$\pm$0.018}  & \underline{6.157$\pm$1.053}  & 6.857$\pm$0.713  & 0.932$\pm$0.016 & 0.865$\pm$0.090 \\
          \midrule 
          SSDiff       & \textbf{2.746$\pm$1.078}  & 6.563$\pm$3.646  & 0.762$\pm$0.184  & 0.663$\pm$0.079  & \textbf{10.817$\pm$1.362} & \textbf{18.063$\pm$1.140} & \textbf{0.737$\pm$0.077} & \textbf{0.785$\pm$0.131} \\
          our.SSDiff   & 1.720$\pm$0.207  & 1.486$\pm$0.448  & 0.958$\pm$0.028  & 0.822$\pm$0.064  & 6.489$\pm$1.263  & 6.043$\pm$0.625  & 0.943$\pm$0.007 & 0.785$\pm$0.130 \\
          \midrule 
          ADWM         & \underline{3.177$\pm$0.651}  & \underline{2.869$\pm$0.807}  & \textbf{0.914$\pm$0.0380}  & \underline{0.784$\pm$0.052}  & 23.593$\pm$11.434 & 25.958$\pm$10.122 & 0.433$\pm$0.141 & 0.169$\pm$0.177 \\
          our.ADWM     & 2.020$\pm$0.364  & 1.768$\pm$0.275  & 0.946$\pm$0.012  & 0.886$\pm$0.024  & 6.346$\pm$1.150  & \underline{5.662$\pm$0.605}  & \underline{0.952$\pm$0.018} & \underline{0.895$\pm$0.102}\\
          \midrule 
          WFANet       & 3.248$\pm$0.452  & 3.033$\pm$0.493  & 0.902$\pm$0.017  & 0.727$\pm$0.070  & 50.510$\pm$6.395 & 24.903$\pm$1.083 & 0.753$\pm$0.056 & 0.384$\pm$0.113 \\
          our.WFANet   & \textbf{0.978$\pm$0.168}  & \textbf{0.894$\pm$0.130} & \textbf{0.982$\pm$0.003}  & \textbf{0.963$\pm$0.018} & \textbf{4.778$\pm$0.776}  & \textbf{3.946$\pm$0.291} & \textbf{0.980$\pm$0.007} & \textbf{0.930$\pm$0.090} \\
      \bottomrule
      \bottomrule
    \end{tabular}
  \caption{Mean values and standard deviations of all comparative methods on 20 reduced-resolution samples from the QB and GF2 datasets. Best:\textbf{bold} , and second best: :\underline{underline}.}
  \label{tab:qb_gf2_reduced_results}
\end{minipage}
\end{table*}

\textbf{Benchmark:}
To evaluate the effectiveness of the SWIFT framework, we compare it with a series of representative deep learning models in the field of pansharpening, covering architectures from classical to the latest ones, including PanNet \cite{yang2017}, FusionNet \cite{deng2021}, U2Net \cite{peng2023}, SSDiff \cite{yu2024}, ADWM \cite{huang2025general}, and WFANet \cite{Huang2025}. For fairness, all baseline models are pre-trained using the same source-domain dataset as our method, and their hyperparameter settings strictly follow the configurations in their respective original papers. In addition, to further verify the advancement of our method, we provide performance comparisons of more models on non-cross-domain tasks in the supplementary materials to confirm our SOTA performance.
 
\begin{figure*}
    \centering
    \includegraphics[width=1\linewidth]{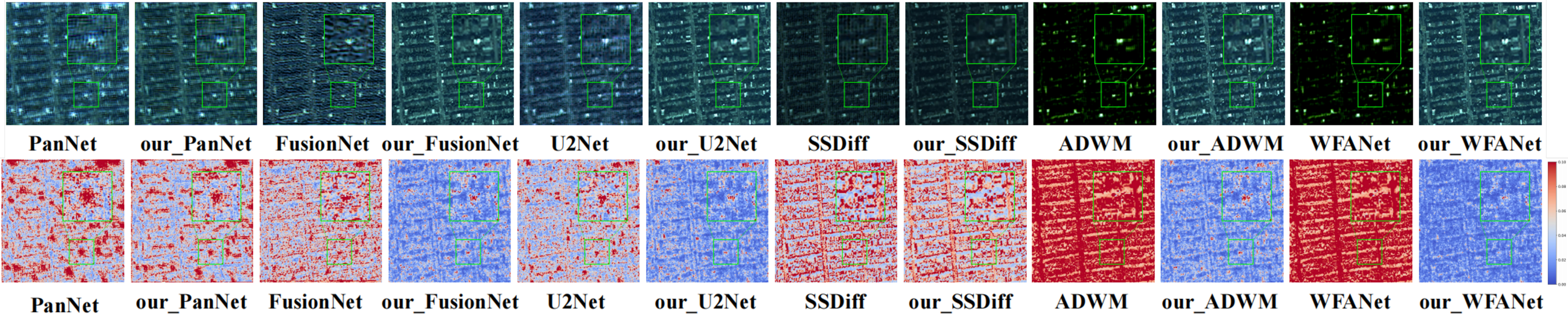}
    \caption{Visual Fusion image and Error maps on QB dataset (reduced data). For the error maps, blue indicate low error.}
    \label{fig: error maps}
\end{figure*}

\subsection{Main Experimental Results}

Our main experimental results, presented in Tables \ref{tab:qb_gf2_full_results}, \ref{tab:qb_gf2_reduced_results}, \ref{tab:wv3_wv2_full_results} and Figures 4-5, demonstrate that the SWIFT framework significantly enhances the cross-sensor performance of various pansharpening models. Qualitatively, the visual comparisons in Figure 3 show that SWIFT-enhanced networks consistently produce results with corresponding error maps exhibiting significantly smaller residuals.Notably, these results also validate our initial claim from the introduction: as shown in our cross-domain experiments (e.g., Table \ref{tab:qb_gf2_reduced_results}), the classic FusionNet model consistently outperforms the more recent ADWM which is designed above Fusionnet, confirming that increasing architectural complexity does not guarantee better generalization.

Quantitatively, the improvements are substantial across all tested datasets. For instance, when applied to WFANet, SWIFT decreases the ERGAS metric by a remarkable 20.957 on the reduced-resolution QB data (Table \ref{tab:qb_gf2_reduced_results}). Similarly, for the ADWM baseline, it boosts the HQNR metric by 0.253 on the full-resolution QB data (Table \ref{tab:qb_gf2_full_results}). 

\begin{table}[htbp]
  \small
  \centering 
  \setlength{\tabcolsep}{2pt}
  \renewcommand\arraystretch{1.1}
    \begin{tabular}{
      p{1.80cm}@{} 
      >{\centering\arraybackslash}p{1.69cm} 
      >{\centering\arraybackslash}p{1.69cm} 
      >{\centering\arraybackslash}p{1.75cm} 
      >{\centering\arraybackslash}p{0.86cm}
    }
      \toprule
      \toprule
      \multirow{2}{*}{Method} & \multicolumn{4}{c}{WV3-WV2: Avg$\pm$std}  \\
      \cmidrule{2-5}
       & $D_\lambda\downarrow$ & $D_s\downarrow$ & HQNR$\uparrow$ & Time(s) \\
      \midrule
          FusionNet & \textbf{0.054$\pm$0.027} & 0.059$\pm$0.015 & \underline{0.889$\pm$0.022} & \textbf{2630} \\
          our.FusionNet & \textbf{0.034$\pm$0.015} & 0.039$\pm$0.009 & 0.928$\pm$0.011 & \textbf{78.0} \\
          \midrule 
          U2Net & 0.094$\pm$0.079 & \underline{0.004$\pm$0.010} & 0.871$\pm$0.081 & 35129 \\
          our.U2Net & 0.039$\pm$0.019 & \textbf{0.021$\pm$0.006} & \textbf{0.941$\pm$0.022} & 243.2 \\
          \midrule 
          WFANet & \underline{0.063$\pm$0.040} & \textbf{0.033$\pm$0.006} & \textbf{0.907 $\pm$0.042} & \underline{16481} \\
          our.WFANet & \underline{0.035$\pm$0.015}  & \underline{0.030$\pm$0.005} & \underline{0.937$\pm$0.018}  & \underline{203.7} \\
      \bottomrule
      \bottomrule
    \end{tabular}
  \caption{Mean values and standard deviations of all comparative methods on 20 full-resolution samples from the WV2 datasets. Best:\textbf{bold} , and second best: :\underline{underline}.}
  \label{tab:wv3_wv2_full_results}
\end{table}

\begin{figure}[h]
    \centering
    \includegraphics[width=0.9\linewidth]{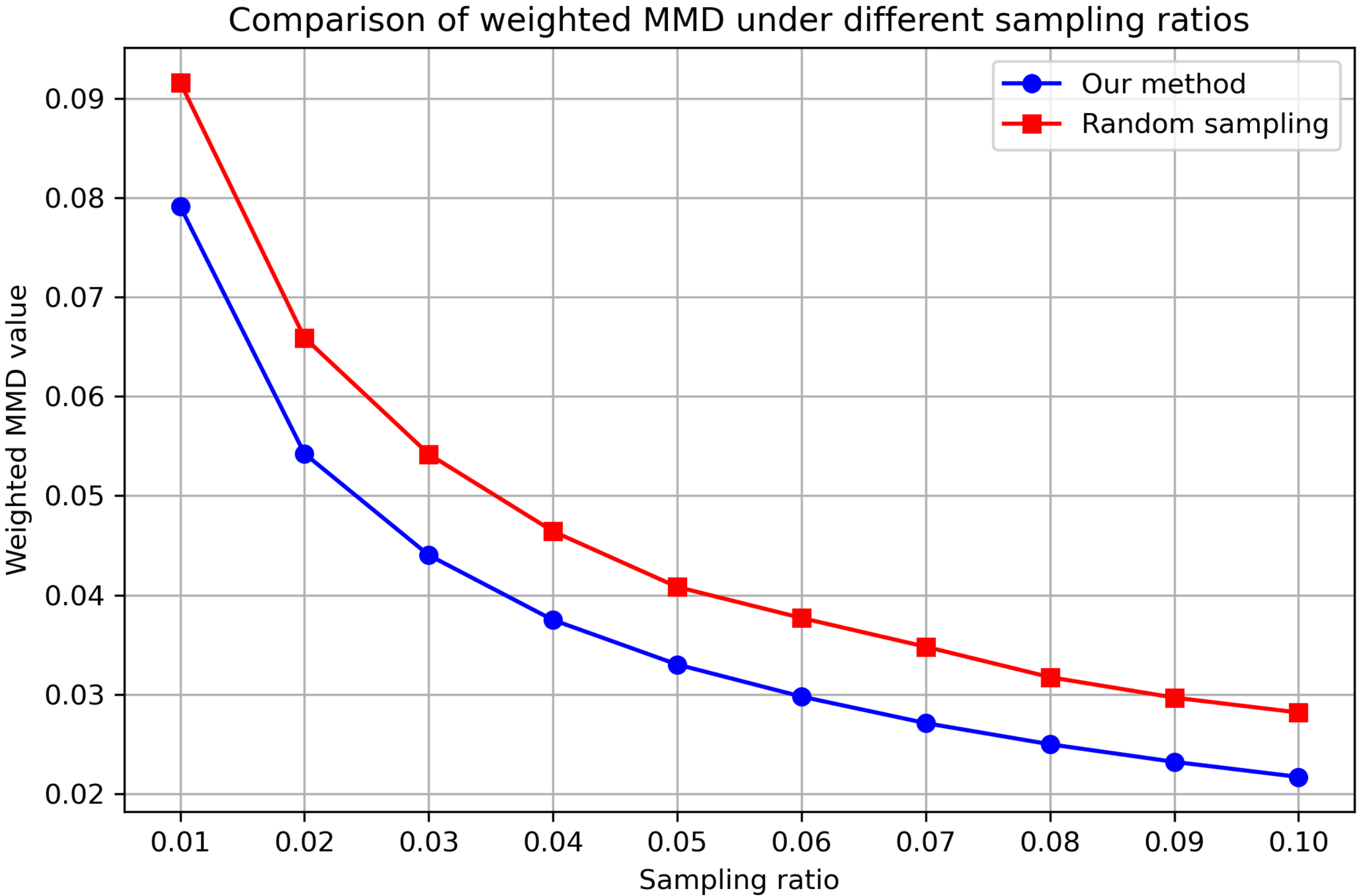}
    \caption{Comparison of Weighted MMD (Maximum Mean Discrepancy) between our proposed sampling method and random sampling under different sampling ratios. }
    \label{fig:mmd}
\end{figure}

These significant performance gains are achieved with remarkable efficiency. \textit{It is crucial to note that} the reported Time(s) metric signifies different procedures: for baseline models (e.g., PanNet), it is the full retraining time on the target domain, whereas for our SWIFT-enhanced models (e.g., our.PanNet), it denotes the rapid adaptation time.This comparison is fair and reflects a practical application scenario, as all methods start from the same pre-trained source-domain model; we focus on evaluating the marginal cost required for adaptation: full retraining versus our SWIFT enhancement. And SWIFT requires only 3\% of the target-domain samples and adapts approximately 30\% of the model's parameters. More notably, the adaptation time for a model like SSDiff is reduced from nearly 10 hours to just under 20 minutes—an approximately 30-fold improvement in efficiency (Table \ref{tab:qb_gf2_full_results}). This series of results validates the effectiveness of SWIFT and highlights its comprehensive advantages in balancing performance with data, parameter, and time efficiency.

\subsection{Ablation Study}

To verify the effectiveness of the "essence" sample selection strategy and the key parameter identification and update mechanism, we designed a series of ablation experiments on the QB dataset using the FusionNet baseline model: Figure N presents the ablation results comparing the dynamic parameter selection strategy based on parameter sharpness with the random selection strategy, where the selection ratio increases from 10\% to 100\% in 10\% increments, within the parameter update mechanism.

\begin{table}[htbp]
  \small
  \centering 
  \setlength{\tabcolsep}{2pt}
  \renewcommand\arraystretch{1.1}
    \begin{tabular}{
      p{1.25cm}@{} 
      >{\centering\arraybackslash}p{1.8cm} 
      >{\centering\arraybackslash}p{1.8cm} 
      >{\centering\arraybackslash}p{1.8cm} 
      >{\centering\arraybackslash}p{0.91cm}
    }
      \toprule
      \toprule
      \multirow{2}{*}{Method} & \multicolumn{4}{c}{GF2-QB-Full-data: Avg$\pm$std} \\
      \cmidrule{2-5}
       & $D_\lambda\downarrow$ & $D_s\downarrow$ & HQNR$\uparrow$ & Time(s) \\
     \midrule
      DA-FPS & 0.136$\pm$0.046 & 0.118$\pm$0.024 & 0.763$\pm$0.058 & 69.3 \\
      \midrule
      Random & 0.138$\pm$0.038 & 0.747$\pm$0.051 & 0.134$\pm$0.029 & 61.5 \\
      \bottomrule
      \bottomrule
    \end{tabular}
  \caption{Comparison of the effects of density - aware farthest point sampling and random sampling at a 3\% ratio on reduced - resolution samples and full - resolution samples, on the QB dataset, using the FusionNet benchmark model.}
  \label{tab:density-aware_farthest_point_sampling_ablation}
\end{table}

\textbf{Density-Aware Farthest Point Sampling strategy:}
To verify the effectiveness of our DA-FPS strategy, we compared it against random sampling at various sampling ratios. The results, shown in Figure 5 and Table \ref{tab:density-aware_farthest_point_sampling_ablation}, confirm that our density-aware approach is more advantageous in selecting samples with high learning value.

\textbf{Dynamic Parameter Selection:}
To validate our dynamic selection strategy, we compared it against fixed-ratio selection from 10\% to 100\%. The results in Figure N demonstrate that our dynamic method, while retaining only 34.1\% of parameters, surpasses the performance of any fixed-ratio selection and even matches that of full fine-tuning, but at a fraction of the computational cost.
\begin{figure}[h]
    \centering
    \includegraphics[width=0.9\linewidth]{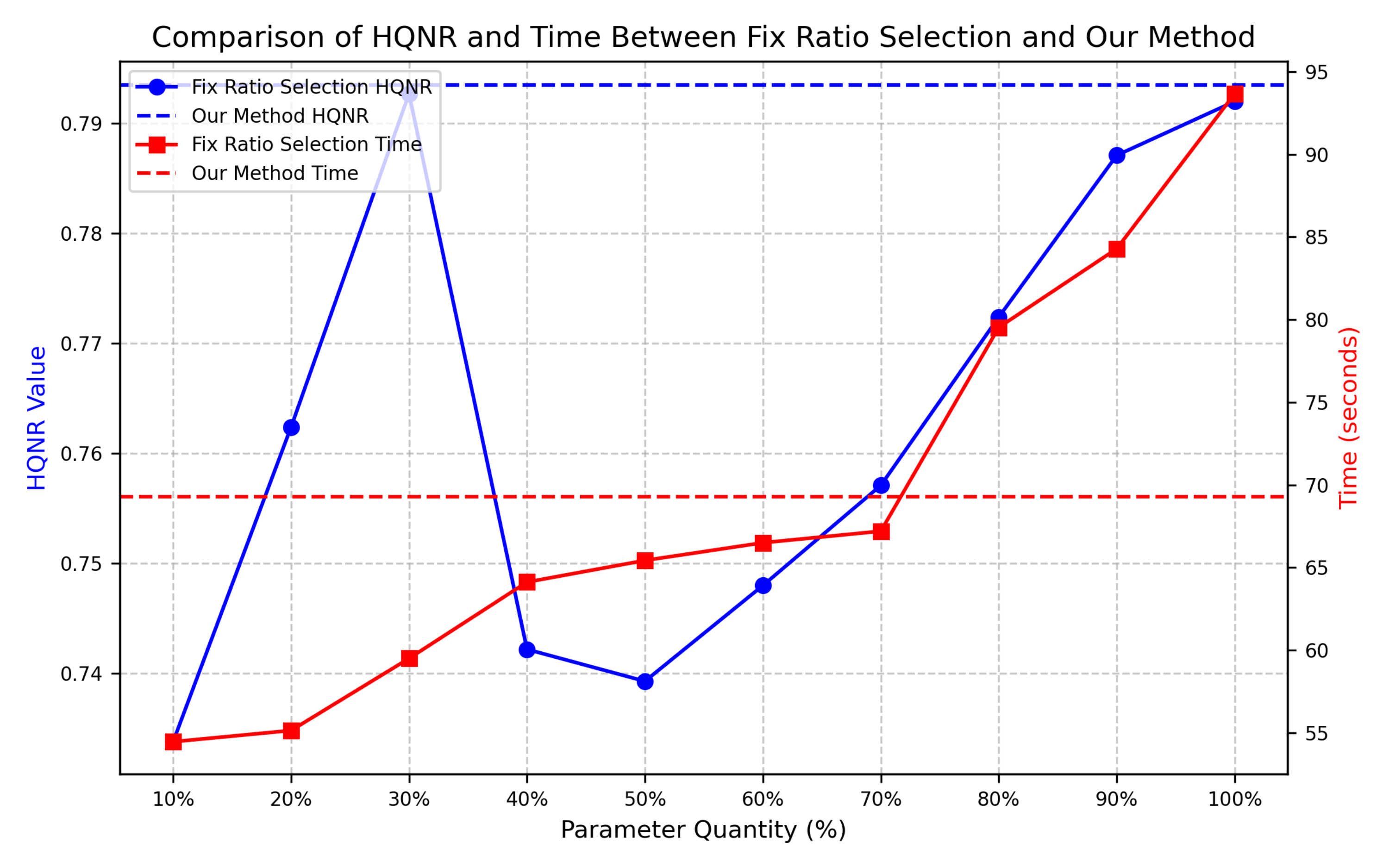}
    \caption{ Performance and efficiency comparison of our dynamic parameter selection strategy against fixed-ratio selection. The blue line (left y-axis) shows the HQNR score, while the red line (right y-axis) shows the adaptation time.}
    \label{fig:enter-label}
\end{figure}
\section{Conclusion}

This paper introduced SWIFT, a model-agnostic enhancement framework designed to address the novel task of improving cross-sensor generalization at an extremely low cost without altering the model architecture. Extensive experiments demonstrate that SWIFT achieves state-of-the-art (SOTA) performance on the HQNR metric across the WV2 and QB datasets, often surpassing full retraining while using only a fraction of the data (3\%) and tunable parameters (\~30\%). By efficiently balancing high performance with minimal adaptation costs, SWIFT offers a new and practical paradigm for deploying pansharpening models across diverse sensor domains.

\bibliography{aaai2026}

@article{Crampton2013,
  author    = {Jeremy W. Crampton and Mark Graham and Ate Poorthuis and Taylor Shelton and Monica Stephens and Matthew W. Wilson and Matthew Zook},
  title     = {Beyond the geotag: Situating ‘big data’ and leveraging the potential of the geoweb},
  journal   = {Cartography and Geographic Information Science},
  volume    = {40},
  number    = {2},
  pages     = {130--139},
  year      = {2013},
  month     = mar,
  publisher = {Informa UK Limited},
  doi       = {10.1080/15230406.2013.777137}
}

@article{Fitzner2013,
  author    = {Daniel Fitzner and Monika Sester and Uwe Haberlandt and Ehsan Rabiei},
  title     = {Rainfall estimation with a geosensor network of cars: Theoretical considerations and first results},
  journal   = {Photogrammetrie, Fernerkundung, Geoinformation},
  volume    = {2013},
  number    = {2},
  pages     = {93--103},
  year      = {2013},
  doi       = {10.1127/1432-8364/2013/0161}
}

@article{choi2011,
  author={Myungjin Choi and Rae Young Kim and Myeong-Ryong Nam and Hong Oh Kim},
  journal={IEEE Geoscience and Remote Sensing Letters}, 
  title={Fusion of multispectral and panchromatic Satellite images using the curvelet transform}, 
  year={2005},
  volume={2},
  number={2},
  pages={136-140},
  doi={10.1109/LGRS.2005.845313}
}

@article{vivone2019,
  author={Vivone, Gemine and Alparone, Luciano and Chanussot, Jocelyn and Dalla Mura, Mauro and Garzelli, Andrea and Licciardi, Giorgio A. and Restaino, Rocco and Wald, Lucien},
  journal={IEEE Transactions on Geoscience and Remote Sensing}, 
  title={A Critical Comparison Among Pansharpening Algorithms}, 
  year={2015},
  volume={53},
  number={5},
  pages={2565-2586},
  doi={10.1109/TGRS.2014.2361734}
}

@article{otazu2005,
  author={Otazu, X. and Gonzalez-Audicana, M. and Fors, O. and Nunez, J.},
  journal={IEEE Transactions on Geoscience and Remote Sensing}, 
  title={Introduction of sensor spectral response into image fusion methods. Application to wavelet-based methods}, 
  year={2005},
  volume={43},
  number={10},
  pages={2376-2385},
  doi={10.1109/TGRS.2005.856106}
}

@article{vivone2018,
  author={Vivone, Gemine and Restaino, Rocco and Chanussot, Jocelyn},
  journal={IEEE Transactions on Geoscience and Remote Sensing}, 
  title={A Regression-Based High-Pass Modulation Pansharpening Approach}, 
  year={2018},
  volume={56},
  number={2},
  pages={984-996},
  doi={10.1109/TGRS.2017.2757508}
}

@inproceedings{wu2025,
  title={Panadapter: Two-stage fine-tuning with spatial-spectral priors injecting for pansharpening},
  author={Wu, RuoCheng and Zhang, ZiEn and Deng, ShangQi and Duan, YuLe and Deng, Liang-Jian},
  booktitle={Proceedings of the AAAI Conference on Artificial Intelligence},
  volume={39},
  number={8},
  pages={8450--8459},
  year={2025}
}

@article{tian2021,
  author={Tian, Xin and Chen, Yuerong and Yang, Changcai and Ma, Jiayi},
  journal={IEEE Transactions on Geoscience and Remote Sensing}, 
  title={Variational Pansharpening by Exploiting Cartoon-Texture Similarities}, 
  year={2022},
  volume={60},
  number={},
  pages={1-16},
  doi={10.1109/TGRS.2020.3048257}
}

@article{masi2016,
  AUTHOR = {Masi, Giuseppe and Cozzolino, Davide and Verdoliva, Luisa and Scarpa, Giuseppe},
  TITLE = {Pansharpening by Convolutional Neural Networks},
  JOURNAL = {Remote Sensing},
  VOLUME = {8},
  YEAR = {2016},
  NUMBER = {7},
  ARTICLE-NUMBER = {594},
  URL = {https://www.mdpi.com/2072-4292/8/7/594},
  ISSN = {2072-4292},
  DOI = {10.3390/rs8070594}

}

@article{he2019,
  title={A deep spatial-spectral network for pansharpening},
  author={He, Yuming and Zhong, Yanfei and Ma, Ailong},
  journal={Remote Sensing},
  volume={11},
  number={17},
  pages={2061},
  year={2019},
  publisher={MDPI}
}

@InProceedings{Yang2023,
    author    = {Yang, Gang and Cao, Xiangyong and Xiao, Wenzhe and Zhou, Man and Liu, Aiping and Chen, Xun and Meng, Deyu},
    title     = {PanFlowNet: A Flow-Based Deep Network for Pan-Sharpening},
    booktitle = {Proceedings of the IEEE/CVF International Conference on Computer Vision (ICCV)},
    month     = {October},
    year      = {2023},
    pages     = {16857-16867}
}

@article{Li2024,
  author = {Z. Li and H. Chen and J. Li and others},
  title = {FusFormer: Global and detail feature fusion transformer for semantic segmentation of small objects},
  journal = {Multimedia Tools and Applications},
  volume = {83},
  pages = {88717--88744},
  year = {2024},
  doi = {10.1007/s11042-024-18911-8},
  url = {https://doi.org/10.1007/s11042-024-18911-8}
}

@article{zhang2021,
    title = {GTP-PNet: A residual learning network based on gradient transformation prior for pansharpening},
    journal = {ISPRS Journal of Photogrammetry and Remote Sensing},
    volume = {172},
    pages = {223-239},
    year = {2021},
    issn = {0924-2716},
    doi = {https://doi.org/10.1016/j.isprsjprs.2020.12.014},
    url = {https://www.sciencedirect.com/science/article/pii/S092427162030352X},
    author = {Hao Zhang and Jiayi Ma}

}

@inproceedings{zhou2022,
  author={Zhou, Huanyu and Liu, Qingjie and Wang, Yunhong},
  booktitle={2022 IEEE International Conference on Multimedia and Expo (ICME)}, 
  title={PanFormer: A Transformer Based Model for Pan-Sharpening}, 
  year={2022},
  volume={},
  number={},
  pages={1-6},
  doi={10.1109/ICME52920.2022.9859770}
}

@article{meng2023,
  author={Meng, Qingyan and Shi, Wenxu and Li, Sijia and Zhang, Linlin},
  journal={IEEE Transactions on Geoscience and Remote Sensing}, 
  title={PanDiff: A Novel Pansharpening Method Based on Denoising Diffusion Probabilistic Model}, 
  year={2023},
  volume={61},
  number={},
  pages={1-17},
  doi={10.1109/TGRS.2023.3279864}
}

@article{zhong2024,
  title={Ssdiff: Spatial-spectral integrated diffusion model for remote sensing pansharpening},
  author={Zhong, Yu and Wu, Xiao and Cao, Zihan and Dou, Hong-Xia and Deng, Liang-Jian},
  journal={Advances in Neural Information Processing Systems},
  volume={37},
  pages={77962--77986},
  year={2024}
}

@article{rui2024,
    title = {Unsupervised hyperspectral pansharpening via low-rank diffusion model},
    journal = {Information Fusion},
    volume = {107},
    pages = {102325},
    year = {2024},
    issn = {1566-2535},
    doi = {https://doi.org/10.1016/j.inffus.2024.102325},
    url = {https://www.sciencedirect.com/science/article/pii/S1566253524001039},
    author = {Xiangyu Rui and Xiangyong Cao and Li Pang and Zeyu Zhu and Zongsheng Yue and Deyu Meng}
    
}

@article{quan2021fusionnet,
  title={Fusionnet: A deep fully residual convolutional neural network for image segmentation in connectomics},
  author={Quan, Tran Minh and Hildebrand, David Grant Colburn and Jeong, Won-Ki},
  journal={Frontiers in Computer Science},
  volume={3},
  pages={613981},
  year={2021},
  publisher={Frontiers Media SA}
}

@inproceedings{huang2025general,
  title={A General Adaptive Dual-level Weighting Mechanism for Remote Sensing Pansharpening},
  author={Huang, Jie and Chen, Haorui and Ren, Jiaxuan and Peng, Siran and Deng, Liangjian},
  booktitle={Proceedings of the Computer Vision and Pattern Recognition Conference},
  pages={7447--7456},
  year={2025}
}

@inproceedings{LeCun1989,
  author    = {Yann Le Cun and John S. Denker and Sara A. Solla},
  title     = {Optimal Brain Damage},
  booktitle = {NIPS'89: Proceedings of the 3rd International Conference on Neural Information Processing Systems},
  pages     = {598--605},
  year      = {1989},
  publisher = {Association for Computing Machinery},
}

@INPROCEEDINGS{yang2017,
  author={Yang, Junfeng and Fu, Xueyang and Hu, Yuwen and Huang, Yue and Ding, Xinghao and Paisley, John},
  booktitle={2017 IEEE International Conference on Computer Vision (ICCV)}, 
  title={PanNet: A Deep Network Architecture for Pan-Sharpening}, 
  year={2017},
  volume={},
  number={},
  pages={1753-1761},
  doi={10.1109/ICCV.2017.193}}

@ARTICLE{deng2021,
  author={Deng, Liang-Jian and Vivone, Gemine and Jin, Cheng and Chanussot, Jocelyn},
  journal={IEEE Transactions on Geoscience and Remote Sensing}, 
  title={Detail Injection-Based Deep Convolutional Neural Networks for Pansharpening}, 
  year={2021},
  volume={59},
  number={8},
  pages={6995-7010},
  doi={10.1109/TGRS.2020.3031366}}

@inproceedings{Yuhas1992,
  title={Discrimination among semi-arid landscape endmembers using the Spectral Angle Mapper (SAM) algorithm},
  author={Roberta H. Yuhas and Alexander F. H. Goetz and Joseph W. Boardman},
  booktitle={Summaries of the Third Annual JPL Airborne Geoscience Workshop. Volume 1: AVIRIS Workshop},
  year={1992},
  url={https://api.semanticscholar.org/CorpusID:126879175}
}

@book{wald2002,
  title={Data Fusion: Definitions and Architectures : Fusion of Images of Different Spatial Resolutions},
  author={Wald, L.},
  isbn={9782911762383},
  url={https://books.google.com.sg/books?id=JOQ6elyzq-8C},
  year={2002},
  publisher={Presses de l'{\'E}cole des Mines}
}

@article{zhou1998,
  title={A wavelet transform method to merge Landsat TM and SPOT panchromatic data},
  author={J. T. Zhou and Daniel L. Civco and John A. Silander},
  journal={International Journal of Remote Sensing},
  year={1998},
  volume={19},
  pages={743-757},
  url={https://api.semanticscholar.org/CorpusID:129860303}
}

@ARTICLE{5159503,
  author={Garzelli, Andrea and Nencini, Filippo},
  journal={IEEE Geoscience and Remote Sensing Letters}, 
  title={Hypercomplex Quality Assessment of Multi/Hyperspectral Images}, 
  year={2009},
  volume={6},
  number={4},
  pages={662-665},
  doi={10.1109/LGRS.2009.2022650}}

@ARTICLE{Vivone2015,
  author={Vivone, Gemine and Alparone, Luciano and Chanussot, Jocelyn and Dalla Mura, Mauro and Garzelli, Andrea and Licciardi, Giorgio A. and Restaino, Rocco and Wald, Lucien},
  journal={IEEE Transactions on Geoscience and Remote Sensing}, 
  title={A Critical Comparison Among Pansharpening Algorithms}, 
  year={2015},
  volume={53},
  number={5},
  pages={2565-2586},
  doi={10.1109/TGRS.2014.2361734}}

@inproceedings{peng2023,
author = {Peng, Siran and Guo, Chenhao and Wu, Xiao and Deng, Liang-Jian},
title = {U2Net: A General Framework with Spatial-Spectral-Integrated Double U-Net for Image Fusion},
year = {2023},
isbn = {9798400701085},
publisher = {Association for Computing Machinery},
address = {New York, NY, USA},
url = {https://doi.org/10.1145/3581783.3612084},
doi = {10.1145/3581783.3612084},
booktitle = {Proceedings of the 31st ACM International Conference on Multimedia},
pages = {3219–3227},
numpages = {9},
location = {Ottawa ON, Canada},
series = {MM '23}
}

@inproceedings{yu2024,
author = {Zhong, Yu and Wu, Xiao and Cao, Zihan and Dou, Hong-Xia and Deng, Liang-Jian},
title = {SSDiff: spatial-spectral integrated diffusion model for remote sensing pansharpening},
year = {2025},
isbn = {9798331314385},
publisher = {Curran Associates Inc.},
address = {Red Hook, NY, USA},
booktitle = {Proceedings of the 38th International Conference on Neural Information Processing Systems},
articleno = {2478},
numpages = {25},
location = {Vancouver, BC, Canada},
series = {NIPS '24}
}

@misc{Huang2025,
      title={Wavelet-Assisted Multi-Frequency Attention Network for Pansharpening}, 
      author={Jie Huang and Rui Huang and Jinghao Xu and Siran Pen and Yule Duan and Liangjian Deng},
      year={2025},
      eprint={2502.04903},
      archivePrefix={arXiv},
      primaryClass={eess.IV},
      url={https://arxiv.org/abs/2502.04903}, 
}

\end{document}